\pdfoutput=1
\documentclass[11pt]{article}

\usepackage[]{ACL2023}
\usepackage{hyperref}
\usepackage{lipsum}
\usepackage{times}
\usepackage{latexsym}
\usepackage{tabularx}
\usepackage{multirow}
\usepackage[T1]{fontenc}
\usepackage[utf8]{inputenc}
\usepackage{microtype}

\usepackage{inconsolata}
\usepackage{graphicx}
\graphicspath{ {./Images/} }

\title{L3Cube-MahaNews: News-based Short Text and Long Document Classification Datasets in Marathi}
\author{First Author \\
  Affiliation / Address line 1 \\
  Affiliation / Address line 2 \\
  Affiliation / Address line 3 \\
  \texttt{email@domain} \\\And
  Second Author \\
  Affiliation / Address line 1 \\
  Affiliation / Address line 2 \\
  Affiliation / Address line 3 \\
  \texttt{email@domain} \\}

  \author{ Saloni Mittal\textsuperscript{1,3}, Vidula Magdum\textsuperscript{1,3}, Omkar Dhekane\textsuperscript{1,3}, Sharayu Hiwarkhedkar\textsuperscript{1,3} \\
\textbf{Raviraj Joshi}\textsuperscript{2,3} \\
  \textsuperscript{1} Pune Institute of Computer Technology, Pune, Maharashtra India \\ 
  \textsuperscript{2} Indian Institute of Technology Madras, Chennai, Tamil Nadu India\\ 
  \textsuperscript{3} L3Cube Labs, Pune\\
  \texttt{\{salonimittal12, vidulamagdum12, omkarjd1212, hiwarkhedkarsharayu\}@gmail.com} \\
  \texttt{ravirajoshi@gmail.com}}

\begin{document}
\maketitle
\begin{abstract}
%There has been a very low work done for low-resource %languages 
%One of the reasons for this is the substantial amount of time needed for manual annotation work.
%Marathi is the third most spoken language in India and ranks in the top 22 languages spoken worldwide. Despite its popularity, this low-resource language lacks labeled datasets required for machine learning tasks. 

%older

% The currently available text or topic classification datasets in the low-resource Marathi language do not have more than 4 target labels, and some achieve near-perfect accuracy.
% We present L3Cube-MahaNews, a Marathi text classification corpus based on news headlines and articles. It is the largest supervised Marathi Corpus, housing 1.05L+ records classified into a broad spectrum of 12 categories. The MahaNews corpus comprises three supervised datasets for the classification of short text, long documents, and medium paragraphs. The consistent labels across datasets make it suitable for document length-based analysis. We report the data statistics and baseline numbers on these datasets using the pre-trained state-of-the-art BERT models.
% We present a comparative analysis of monolingual and multilingual BERT models such as MahaBERT, IndicBERT, and MuRIL. We report the best results with the monolingual MahaBERT model on all the datasets. The resources can also be viewed as Marathi topic classification datasets or models, and they are publicly shared at \url{https://github.com/l3cube-pune/MarathiNLP}.

%new
The availability of text or topic classification datasets in the low-resource Marathi language is limited, typically consisting of fewer than 4 target labels, with some achieving nearly perfect accuracy. In this work, we introduce L3Cube-MahaNews, a Marathi text classification corpus that focuses on News headlines and articles. This corpus stands out as the largest supervised Marathi Corpus, containing over 1.05L records classified into a diverse range of 12 categories. To accommodate different document lengths, MahaNews comprises three supervised datasets specifically designed for short text, long documents, and medium paragraphs. The consistent labeling across these datasets facilitates document length-based analysis.
We provide detailed data statistics and baseline results on these datasets using state-of-the-art pre-trained BERT models. We conduct a comparative analysis between monolingual and multilingual BERT models, including MahaBERT, IndicBERT, and MuRIL. The monolingual MahaBERT model outperforms all others on every dataset. These resources also serve as Marathi topic classification datasets or models and are publicly available at \url{https://github.com/l3cube-pune/MarathiNLP}.

\end{abstract}
\textbf{Keywords: } Marathi Text Classification, Marathi Topic Identification, Low Resource Language, Short Text Classification, Long Document Classification, News Article Datasets, BERT, Web Scraping.

\section{Introduction}

%Some of the most crucial phases in machine learning are data collection and segregation. Data is vital in NLP for training, categorization, text analysis, etc. 
Text Classification is a popular problem often discussed in machine learning and natural language processing (NLP) \cite{minaee2021deep}. It deals with organizing, segregating, and appropriately assigning the textual sentence or a document into some predefined categories. It is a supervised learning task and has been solved using traditional machine learning approaches and more recent deep learning algorithms \cite{wagh2021comparative}. Text classification is important for applications like the automatic categorization of web articles or social media comments. While a lot of research has been done in the area of English text classification, low-resource languages like Marathi are still left behind. In this work, we focus on the classification of text in the Marathi language.
%Labeled data is required for supervised machine learning. Thus, text classification plays an important role in categorizing data and assigning labels to it. To improve the trained model's accuracy, the ground truth must be equally correct.

The Marathi language is one of the 22+ Indian languages\footnote{\url{https://en.wikipedia.org/wiki/Languages_of_India}} out of the 7000 languages spoken worldwide\footnote{\url{https://en.wikipedia.org/wiki/Lists_of_languages}}. It is the third most spoken language of India, spoken by over 83 million people across the country. It ranks 11th in the list of popular languages across the globe\footnote{\url{https://en.wikipedia.org/wiki/Marathi_language}}. Despite being a widely spoken language, Marathi-specific NLP monolingual resources are still limited in comparison to other natural languages \cite{joshi-2022-l3cube}. As a result, sufficient data resources for machine learning tasks are less available for this language, making it challenging for researchers conducting studies in this widely used though low resource-based regional language. It can be noticed that the datasets available are largely in Mandarin Chinese, Spanish, English, Arabic, Hindi, and Bengali languages\footnote{\url{https://en.wikipedia.org/wiki/List_of_datasets_for_machine-learning_research}}. There are fewer datasets on regional languages like Marathi. The only four classification datasets publicly available are iNLTK headlines \cite{arora2020inltk}, IndicNLP articles \cite{kakwani2020indicnlpsuite}, MahaHate \cite{velankar2022l3cubemahahate}, and MahaSent \cite{kulkarni-etal-2021-l3cubemahasent}. \cite{Kulkarni_2022} showed that the IndicNLP News Article dataset achieves near-perfect accuracy (99\%) thus limiting its usability. Therefore we need some complex datasets to evaluate the goodness of the models. Also, all of these datasets have at most four target labels. Thus, there is a significant need of datasets with exhaustive labels similar to that of BBC News\footnote{\url{https://www.kaggle.com/c/learn-ai-bbc}} or AG News\footnote{\url{https://www.kaggle.com/datasets/amananandrai/ag-news-classification-dataset}} for the language Marathi.
%as there is no purely long document dataset in this language.

Datasets with varying sequence lengths are required as transformer-based classification models are sensitive towards the text length due to their self-attention operation 
\cite{beltagy2020longformer}. Models like LongFormer \cite{beltagy2020longformer} are specifically developed for datasets having longer sequences. In order to develop these models for Marathi we first need such target datasets. Thus, we present \textbf{L3Cube-MahaNews - A Marathi News Classification Dataset} in this paper.

% Creating the correct dataset is critical for machines to learn and train themselves to execute tasks more effectively. Finance, Economics, Commerce, Societal, Health, Academy, Sports, Food, Agriculture, Travel, Geospatial, Political, Consumer, Transport, Logistics, Environmental, Real-Estate, Legal, Entertainment, Energy, Hospitality, and so on are a few of the dataset's knowledge domains.

The dataset we propose is available in three forms, viz. short, medium, and long text classification that is obtained from a renowned Marathi news website. 
%This dataset will help researchers and users in working effectively with little to big sets of data that are ordered precisely and well-labeled. 
This massive corpus of over 1 lakh records will serve as an excellent data source for the comparison of different machine-learning algorithms in low-resource settings. 
It contains information about 12 dynamic categories for diverse disciplines of study. We evaluate different monolingual and multilingual BERT \cite{devlin2018bert} models like MahaBERT \cite{joshi-2022-l3cube}, indicBERT \cite{kakwani2020indicnlpsuite}, and MuRil \cite{khanuja2021muril} and provide baseline results for future studies. The results are evaluated using the validation accuracy, testing Accuracy, F1 score (Macro), recall (Macro), and precision (Macro).

The main contributions of this work are as follows:
\begin{itemize}
    \item We present L3Cube-MahaNews, the first extensive document classification dataset in Marathi with 12 target labels. The dataset will be released publicly.
    \item The corpus consists of three sub-datasets MahaNews-SHC, LPC, and LDC for short, medium, and long documents respectively. We provide three different datasets with varying sentence lengths and the same target labels.
    \item The datasets are benchmarked using state-of-the-art BERT models like MahaBERT, MuRIL, and IndicBERT with MahaBERT giving the best results. We thus present a comparative analysis of these monolingual and multilingual BERT models for Marathi text. The MahaNews-LDC-BERT\footnote{\href{https://huggingface.co/l3cube-pune/marathi-topic-long-doc}{l3cube-pune/marathi-topic-long-doc}}, MahaNews-LPC-BERT\footnote{\href{https://huggingface.co/l3cube-pune/marathi-topic-medium-doc}{l3cube-pune/marathi-topic-medium-doc}}, MahaNews-SHC-BERT\footnote{\href{https://huggingface.co/l3cube-pune/marathi-topic-short-doc}{l3cube-pune/marathi-topic-short-doc}}, and MahaNews-All-BERT\footnote{\href{https://huggingface.co/l3cube-pune/marathi-topic-all-doc}{l3cube-pune/marathi-topic-all-doc}} have been released on Hugging Face.
\end{itemize}
%In the following sections, we explain the results as well as a brief notion and description of this Marathi corpus.
% These instructions are for authors submitting papers to ACL 2023 using \LaTeX. They are not self-contained. All authors must follow the general instructions for *ACL proceedings,\footnote{\url{http://acl-org.github.io/ACLPUB/formatting.html}} as well as guidelines set forth in the ACL 2023 call for papers.\footnote{\url{https://2023.aclweb.org/calls/main_conference/}} This document contains additional instructions for the \LaTeX{} style files.
% The templates include the \LaTeX{} source of this document (\texttt{acl2023.tex}),
% the \LaTeX{} style file used to format it (\texttt{acl2023.sty}),
% an ACL bibliography style (\texttt{acl\_natbib.bst}),
% an example bibliography (\texttt{custom.bib}),
% and the bibliography for the ACL Anthology 
%(\texttt{anthology.bib}).
\section{Related Work}
Text classification is a very popular task in Natural Language Processing. Even though Marathi is a widely spoken language, the lack of proper Marathi datasets that can be used for text classification tasks has restricted the area of research for this language. In this section, we review a few of the publicly available Indian language datasets that are used for the objective task.

\citet{kakwani2020indicnlpsuite} curated large-scale sentence-level monolingual corpora- \textit{IndicCorp} containing 11 major Indian languages. 
%The Corpora is primarily created by collecting text from popular online newspaper websites, magazines, blog posts, etc., and then combining it with \textit{OSCAR} corpus\footnote{\url{https://oscar-project.org/}}. It contains a total of 8.8 billion tokens across all 11 languages and Indian English. Marathi is one of the 11 languages in IndicCorp constituting 34 million \cite{kakwani2020indicnlpsuite} sentences of it. 
\textit{IndicNLP News Article} dataset is part of the IndicNLPSuite\footnote{\url{https://github.com/anoopkunchukuttan/indic_nlp_library}} consists of news articles in Marathi categorized into 3 classes viz. sports, entertainment, and lifestyle. The datasets provided by IndicNLP are used to pre-train the word embedding and multilingual models. 
%Further, The author used these datasets to perform downstream tasks - a) News article topic, b) News headlines topic, and c) Sentiment classification task. 

% \cite{mhaske2022naamapadam} presented \textit{Naamapadam}\footnote{\url{https://ai4bharat.org/naamapadam}}, a publicly available Named Entity Recognition dataset for the 11 major Indian languages that include Assamese, Bengali, Gujarati, Hindi, Kannada, Malayalam, Marathi, Oriya, Punjabi, Tamil, Telugu. It has more than 400k sentences in each language, annotated with at least 100k entities from the three standard categories of entities (Person, Location, and Organisation) for nine of the 11 languages. The author provides IndicNER,
% a multilingual mBERT model fine-tuned on the Naamapadam training set. It is shown that IndicNER outperforms an mBERT model when fine-tuned on available datasets in terms of F1 on the Naamapadam test set.

\citet{arora2020inltk} presented iNLTK\footnote{\url{https://github.com/goru001/inltk}} which is an open-source NLP library containing pre-trained language models and methods for data augmentation, textual similarity, tokenization, word embeddings, etc. 
%The library supports datasets from  13 Indic Languages (including English). These datasets\footnote{\url{https://www.kaggle.com/datasets/disisbig/marathi-wikipedia-articles}} are collected from Wikipedia articles using text extractor tool\footnote{\url{https://github.com/attardi/wikiextractor}} and BeautifulSoup\footnote{\url{https://www.crummy.com/software/BeautifulSoup/}} and are used for pretraining ULMFiT and TransformerXL models. Further, they evaluated the pre-trained ULMFiT for text-classification task using their own \textit{iNLTK headlines corpus} along with publicly available datasets. 
The \textit{iNLTK Headlines Corpus - Marathi\footnote{\url{https://www.kaggle.com/datasets/disisbig/marathi-news-dataset}}} is a Marathi News Classification Dataset provided by iNLTK, containing nearly 12000 news article headlines collected from a Marathi news website. The corpus contains 3 label classes viz. state (62\%) entertainment (27\%) sports (10\%)

\citet{eranpurwala2022comparative} presented a comparative study of Marathi text classification using monolingual and multilingual embeddings. For the experiment, they use  \textit{Marathi news headline dataset} sourced from Kaggle with 9K examples and three label classes - entertainment, state, and sport. The news article headlines were originally collected from a Marathi news website. 
%They benchmarked the results with traditional SVM, Naïve Bayes, random forest algorithms, and complex LSTM and biLSTM algorithms. 
Their study also showed that multilingual embeddings have ~15 percent performance gain compared to traditional monolingual embeddings.

\citet{jain2020indic} evaluated and compared the performance of language models on text classification tasks over 3 Indian languages - Hindi, Bengali, and Telugu. For Hindi, they used \textit{BBC Hindi News Articles} which contains annotated news articles classified into 14 different categories. While  for Bengali and Telugu, they used classification datasets provided by Indic-NLP \cite{kakwani2020indicnlpsuite}. Their result demonstrated that monolingual models perform better for some languages but the improvement attained is marginal at best.

\citet{velankar2022l3cubemahahate} curated \textit{MahaHate}- a tweet-based marathi hate speech detection dataset. The dataset is collected from Twitter and annotated manually. It consists of over 25000 distinct tweets labeled into 4 major classes i.e. hate, offensive, profane, and no. 
%Alternatively, they provide a 2-class (HOF, NOT) variant of the dataset that is used for binary classification. The authors provide tweet collection and annotation policy as well. 
The deep learning models based on CNN, LSTM, and transformers that involved monolingual and multilingual variants of BERT were used for evaluation. 

%The Sentiment Analysis is essentially a text classification problem where a class label denotes the sentiment associated with a given input text. It is used by many social media platforms like Facebook or Twitter. 
\citet{kulkarni-etal-2021-l3cubemahasent} offers first major publicly available Marathi Sentiment Analysis Dataset \textit{L3CubeMahaSent}. The dataset is curated using tweets extracted from various Maharashtrian personalities' Twitter accounts. It consists of ~16,000 distinct tweets classified into three classes - positive, negative, and neutral. The authors performed 2-class and 3-class sentiment analysis on their dataset and evaluated baseline classification results using deep learning models - CNN, LSTM, and ULMFiT.

\citet{Velankar_2022} conducted a comparative study
between monolingual and multilingual BERT models. The standard multilingual models such as mBERT, indicBERT, and xlm-RoBERTa along with monolingual models - MahaBERT, MahaALBERT, and MahaRoBERTa for Marathi \cite{joshi-2022-l3cube} were used in this study. 

\begin{table*}[hbt!]
\centering
\begin{tabular}{|l|l|}
\hline
\textbf{Labels} & \textbf{Description}\\
\hline
Auto & Vehicle launches and their reviews\\
\hline
%\\[-1em]
Bhakti & Horoscope, festivals, spirituality  \\ \hline
%\\[-1em]
Crime &  Crimes and accidents in the country  \\ \hline
%\\[-1em]
Education & Educational institutes and their activities\\ \hline
%\\[-1em]
Fashion &  Fashion events, advertisements of fashion products \\ \hline
%\\[-1em]
Health & Diseases, medicines, and health-related blogs  \\  \hline
%\\[-1em]
International & Happenings around the world  \\ \hline
%\\[-1em]
Manoranjan &  Information related to movies, web series, and so on \\ \hline
%\\[-1em]
Politics & Political incidents in the country \\ \hline
%\\[-1em]
Sports &  Various sports games, awards, sporting events and so on \\ \hline
%\\[-1em]
Tech & Latest technologies, gadgets and their reviews \\ \hline
%\\[-1em]
Travel & Travel tips, Top destinations recommendations, tourism information, et cetera. \\
\hline
\end{tabular}
\caption{Categorical labels for MahaNews datasets}
\label{tab:labels}
\end{table*}

\begin{table*}[htb!]
\centering
  \begin{tabular}{|c|c|c|c|c|c|c|c|c|}
    \hline
    \multirow{2}{*}{Labels} &
      \multicolumn{4}{c|}{\textbf{SHC \& LDC}} & 
      % \multicolumn{3}{c|}{\textbf{}} &
      \multicolumn{4}{c|}{\textbf{LPC}} \\ \cline{2-9}
    & Train & Test & Validation & Total & Train & Test & Validation & Total\\
    \hline
    \textbf{Auto} & 1664	& 209 &	208 & 2081 & 3099 & 388 & 387 & 3874 \\
    
    \hline
    \textbf{Bhakti} & 1386  & 174 & 173 & 1733 & 3664 & 458 & 458 & 4580 \\
    
    \hline
    \textbf{Crime} & 2354 & 295 & 294 & 2943 & 4092 & 512 & 512 & 5116\\
    \hline
     \textbf{Education} & 680  & 86 & 85 & 851 & 1438 & 180 & 180 & 1798\\
    \hline
     \textbf{Fashion} & 1920 & 241 & 240 & 2401 & 874 & 110 & 109 & 1093\\
    \hline
     \textbf{Health} & 1985 & 249 & 248 & 2482 & 6428 & 804 & 803 & 8035\\
    \hline
     \textbf{International} & 2041 & 256 & 255 & 2552 & 4715 & 590 & 589 & 5894\\
    \hline
    \textbf{Manoranjan} & 2986 & 374 & 373 & 3733 & 4825 & 604 & 603 & 6032\\
    \hline
    \textbf{Politics} & 2250 & 282 & 281 & 2813 & 4379 & 548 & 547 & 5474\\
    \hline
    \textbf{Sports} & 1882 & 236 & 235 & 2353 & 5337 & 668 & 667 & 6672\\
    \hline
  \textbf{Tech} & 2111 & 264 & 264 & 2639 & 2049 & 257 & 256 & 2562\\
    \hline
     \textbf{Travel} & 755 & 95 & 94 & 944 & 1970 & 247 & 246 & 2463\\
    \hline
     Total & 22014 & 2761 & 2750 & \textbf{27525} & 42870 & 5366 & 5357 & \textbf{53593}\\
    \hline
    
      \end{tabular}
   \caption{Category-wise distribution of SHC, LDC, LPC datasets into train, test and validation in ratio of 80:10:10.}
  \label{tbl:split}

\end{table*}

% \begin{table*}[htb!]
% \centering
%   \begin{tabular}{|c|c|c|c|c|c|c|}
%     \hline
%     \multirow{2}{*}{Labels} &
%       \multicolumn{3}{c|}{\textbf{SHC \& LDC}} & 
%       % \multicolumn{3}{c|}{\textbf{}} &
%       \multicolumn{3}{c|}{\textbf{LPC}} \\ \cline{2-7}
%     & Train & Test & Validation & Train & Test & Validation \\
%     \hline
%     \textbf{Auto} & 1664	& 209 &	208 & 3099 & 388 & 387 \\
%     \hline
%     \textbf{Bhakti} & 1386  & 174 & 173 & 3664 & 458 & 458 \\
%     \hline
%     \textbf{Crime} & 2354 & 295 & 294 & 4092 & 512 & 512 \\
%     \hline
%      \textbf{Education} & 680  & 86 & 85 & 1438 & 180 & 180 \\
%     \hline
%      \textbf{Fashion} & 1920 & 241 & 240 & 874 & 110 & 109 \\
%     \hline
%      \textbf{Health} & 1985 & 249 & 248 & 6428 & 804 & 803 \\
%     \hline
%      \textbf{International} & 2041 & 256 & 255 & 4715 & 590 & 589 \\
%     \hline
%     \textbf{Manoranjan} & 2986 & 374 & 373 & 4825 & 604 & 603 \\
%     \hline
%     \textbf{Politics} & 2250 & 282 & 281 & 4379 & 548 & 547 \\
%     \hline
%     \textbf{Sports} & 1882 & 236 & 235 & 5337 & 668 & 667 \\
%     \hline
%   \textbf{Tech} & 2111 & 264 & 264 & 2049 & 257 & 256 \\
%     \hline
%      \textbf{Travel} & 755 & 95 & 94 & 1970 & 247 & 246 \\
%     \hline
%       \end{tabular}
%    \caption{Category-wise distribution of SHC, LDC, LPC datasets into train, test and validation in ratio of 80:10:10.}
%   \label{tbl:split}

% \end{table*}

\section{Curating the Dataset}
% The L3Cube-MahaNews is a collection of Short and Long Document Classification corpora. MahaNews consists of three supervised datasets namely, Short Document Classification (SHC), Long Document Classification Level 1 (LDC) and Long Document Classification Level 2 (LPC).
We propose L3Cube-MahaNews which is a collection of datasets for short text and long document classification. The Short Headlines Classification (SHC), Long Document Classification (LDC), and Long Paragraph Classification (LPC) datasets are the three supervised datasets included in MahaNews.
\begin{itemize}
  \item \textit{\textbf{Short Headlines Classification (SHC)}}: This Short Document Classification dataset contains the headlines of news articles along with their corresponding categorical labels.
  \item \textit{\textbf{Long Paragraph Classification (LPC)}}: This is a Long Document Classification dataset. The news articles are divided into paragraphs and each record in this dataset contains a paragraph each with its corresponding categorical label.
  \item \textit{\textbf{Long Document Classification (LDC)}}: This Long Document Classification dataset contains records having an entire news article along with its corresponding categorical label.
  
\end{itemize}
The categorical labels in the supervised datasets are described in detail in Table~\ref{tab:labels}.

\begin{figure*}[hbt]
    \centering
   
    \begin{minipage}{0.48\linewidth}
        \centering
        \includegraphics[scale=0.3]{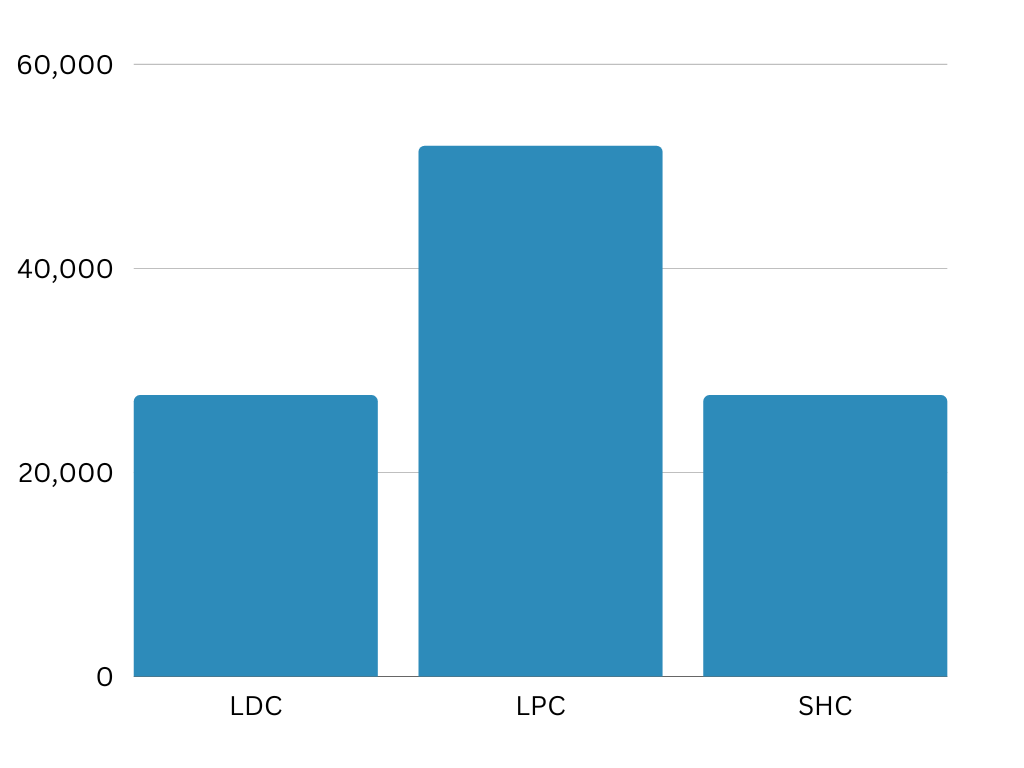}%{c.PNG} % first figure itself
        \caption{Statistical count of records in SHC, LDC, and LPC}
        \label{fig:statistics}
    \end{minipage}\hfill
     \hfill %
    \begin{minipage}{0.48\linewidth}
        \centering
        \includegraphics[scale=0.3]{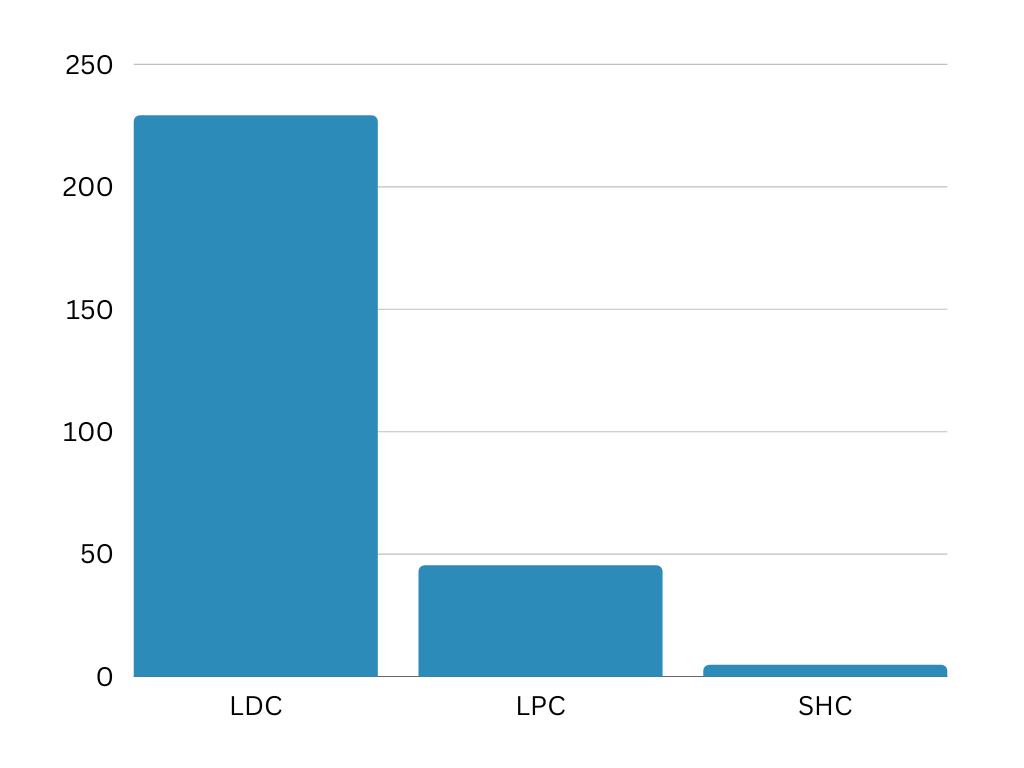}%{h.PNG} % second figure itself
        \caption{Average count of words per record in SHC, LDC, and LPC}
        \label{fig:words}
    \end{minipage}
\end{figure*}
\begin{figure*}[hbt]
    \centering
   
    \begin{minipage}{0.48\linewidth}
        \centering
        \includegraphics[scale=0.9]{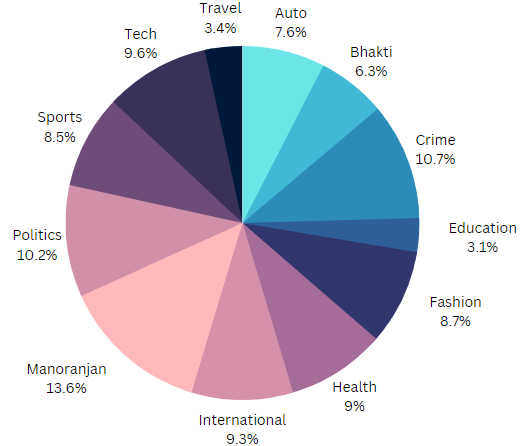}%{c.PNG} % first figure itself
        \caption{Percentage Distribution of 27,525 records of SHC and LDC based on categorical labels}
        \label{fig:LDC}
    \end{minipage}\hfill
     \hfill%
    \begin{minipage}{0.48\linewidth}
        \centering
        \includegraphics[scale=0.9]{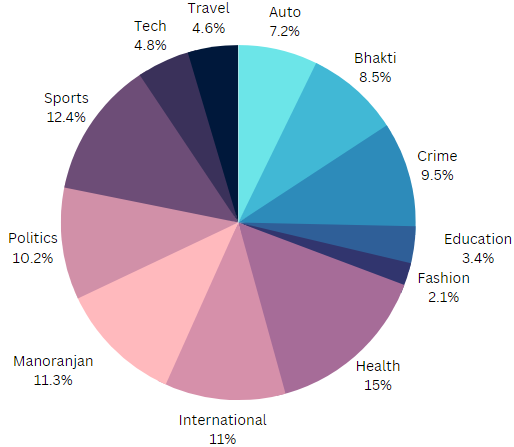}%{h.PNG} % second figure itself
        \caption{Percentage Distribution of 53,593 records of LPC based on categorical labels}
        \label{fig:LPC}
    \end{minipage}
\end{figure*}

\subsection{Data Collection}
The datasets are compiled using scraped news data. The entirety of the information is taken from the Lokmat\footnote{\url{https://www.lokmat.com/}} website which houses news articles in the Marathi language. The data was scraped by using urllib package to handle URL requests and the BeautifulSoup package to extract data from the HTML of the requested URL. 

Lokmat website had arranged the news articles under predefined categories like automobile, sports, travel, politics, etc. While scraping, this categorization was preserved and further used as target labels. The final curated datasets were shuffled, de-duplicated, and cleaned up. 

\subsection{Data Statistics}

% \includegraphics[scale=0.7]{Bar Chart.png}
% \begin{figure*}[]
%   \centering
% \includegraphics[scale=0.35]{BarChart1.png}
% \caption{Statistical count of records in SHC, LDC and LPC}
% \label{fig:overview}
% \end{figure*}
The L3Cube-MahaNews has a total of 1,08,643 records which are derived from 27,525 news articles scraped from Lokmat. SHC and LDC have a total of 27,525 rows with labels each and LPC has 53,593 labeled rows in it. 

The statistical count of records in SHC, LDC, and LPC can be referred from Figure~\ref{fig:statistics} and the average count of words per record in each proposed dataset can be seen in Figure~\ref{fig:words}. 

The category-wise percentage distribution for the corpora can be referred from Figure~\ref{fig:LDC} and~\ref{fig:LPC}.
% \\
% \\
% For more clarity, some example sentences with labels are mentioned in Table 1.
% \begin{table*}
% \centering
% \begin{tabular}{lll}
% \hline
% \textbf{Source} & \textbf{Sentence} & \textbf{Categorical Label}\\
% \hline
% SHC & {Diwali 2022 : \hindifont दिवाळीची खरी सुरुवात वसुबारसेपासूनच; काय आहे गोवत्स द्वादशीचं महत्त्व? जाणून घ्या!} & \verb|\cite| \\
% LDC & {\hindifont नवी दिल्ली : ओला इलेक्ट्रिक स्कूटर लाँच झाल्यापासून भारतीय बाजारपेठेत खूप लोकप्रिय झाली आहे.दरम्यान, कंपनीने अलीकडेच काही शहरांमध्ये या स्कूटरची टेस्ट राइड सुरू केली आहे.आत्तापर्यंत, कंपनीने 1,000 हून अधिक शहरांमध्ये आपल्या इलेक्ट्रिक स्कूटरच्या टेस्ट राइड सुरू करण्याची योजना आखली आहे.ओला इलेक्ट्रिकने एका निवेदनात म्हटले आहे की, सुरुवातीच्या टप्प्यात ज्यांनी कंपनीच्या ओला S1 आणि S1 pro स्कूटर खरेदी केल्या आहेत किंवा रिझर्व्ह केल्या आहेत, त्यांच्यासाठीच टेस्ट राइड खुली असणार आहे.सॉफ्टबँक ग्रुप कॉर्पोरेशन आणि टायगर ग्लोबल मॅनेजमेंटद्वारे समर्थित असलेल्या या फर्मने डिसेंबरपर्यंत सर्व ग्राहकांसाठी टेस्ट राइड सुरू करण्याचे उद्दिष्ट ठेवले आहे......}& no equivalent \\
% LPC & \verb|\citet| & \verb|\newcite| \\

% \hline
% \end{tabular}
% \caption{\label{citation-guide}
% Citation commands supported by the style file.
% The style is based on the natbib package and supports all natbib citation commands.
% It also supports commands defined in previous ACL style files for compatibility.
% }
% \end{table*}

\begin{table*}[hbtp]
  \centering
  \begin{tabular}{|c|c|c|c|c|c|c|}
    \hline

     \multicolumn{2}{|c|}{}  & \begin{tabular}[x]{@{}c@{}} \textbf{Validation} \\\textbf{Accuracy} \end{tabular} & \begin{tabular}[x]{@{}c@{}}\textbf{Testing} \\\textbf{Accuracy}\end{tabular} & \begin{tabular}[x]{@{}c@{}}\textbf{F1 Score} \\\textbf{(Macro)} \end{tabular} & \begin{tabular}[x]{@{}c@{}}\textbf{Recall} \\\textbf{(Macro)}\end{tabular} & \begin{tabular}[x]{@{}c@{}}\textbf{Precision} \\\textbf{(Macro)}\end{tabular} \\\hline
  \multirow{3}{*}{\rotatebox{90}{\textbf{SHC}}}  & MahaBERT & \textbf{91.418} &  \textbf{91.163} &  \textbf{90.230}  &  \textbf{89.700}  &  \textbf{91.047}  \\ %\cline{2-7}    
    & indicBERT & 90.073 &  89.388 &  88.303 &  87.953 &  88.758 \\ %\cline{2-7}
   & MuRIL & 90.655 &  90.112 &  89.031 &  88.826 &  89.313 \\ %\cline{2-7}
   
   \hline  
   \multirow{3}{*}{\rotatebox{90}{\textbf{LDC}}}  & MahaBERT & \textbf{94.780} &  \textbf{94.706} &  \textbf{93.589}  &  \textbf{93.210}  &  \textbf{94.079}  \\ %\cline{2-7}    
    & indicBERT & 93.642 &  92.627 &  91.340 &  91.217 &  91.511 \\ %\cline{2-7}
   & MuRIL &  93.564 &  93.020 &  92.337 &92.213 &  92.501  \\%\cline{2-7}
\hline 
   \multirow{3}{*}{\rotatebox{90}{\textbf{LPC}}}  & MahaBERT & \textbf{88.754} &  \textbf{86.731} &  \textbf{84.915 } &  \textbf{83.455}  &  \textbf{87.138}  \\ %\cline{2-7}    
    & indicBERT & 86.298 &  85.222 &  86.688 &  81.697 &  84.249 \\ %\cline{2-7}
   & MuRIL & 87.157 &  86.582 &  84.585 &  83.215 &  86.603 \\%\cline{2-7}

   \hline  
  \end{tabular}
  \caption{Results for all the models trained on SHC, LDC, and LPC datasets in percentage (\%)}
  \label{tbl:result}
\end{table*}

\begin{table*}[hbt!]
  \centering
  \begin{tabular}{|c|c|c|c|c|c|}
    \hline

      \begin{tabular}[x]{@{}c@{}} \textbf{MahaBERT model} \\\textbf{trained on} \end{tabular}  & \begin{tabular}[x]{@{}c@{}} \textbf{MahaBERT model} \\\textbf{tested on} \end{tabular} & \begin{tabular}[x]{@{}c@{}}\textbf{Testing} \\\textbf{Accuracy}\end{tabular} & \begin{tabular}[x]{@{}c@{}}\textbf{F1 Score} \\\textbf{(Macro)} \end{tabular} & \begin{tabular}[x]{@{}c@{}}\textbf{Recall} \\\textbf{(Macro)}\end{tabular} & \begin{tabular}[x]{@{}c@{}}\textbf{Precision} \\\textbf{(Macro)}\end{tabular} \\
    \hline

    SHC &  & \textbf{91.163} &  \textbf{90.230}  &  \textbf{89.700}  &  \textbf{91.047}\\ %\cline{2-7}
 LPC &\multirow{3}{*}{{SHC}} &  73.234 &  73.001  &  75.669  &  77.353  \\ 
 LDC & & 74.171 &  79.570  &  76.599  & 79.570 \\ %\cline{2-7}%\cline{2-7} 
 SHC+LPC+LDC & &  86.780 &  85.484  &  86.195  &  87.689\\
    \hline

   SHC & &  73.234 &  73.001  &  75.669  &  77.353  \\ %\cline{2-7}    
   
   LPC & \multirow{3}{*}{{LPC}}  & 86.731 & 84.915 & 83.455 & 87.138\\ %\cline{2-7}
   LDC & & 72.201 & 75.741 & 72.530 & 70.521  \\
   SHC+LPC+LDC & &  \textbf{89.713} &  \textbf{88.421}  &  \textbf{88.545}  & \textbf{88.439}\\
\hline
SHC & & 80.314 & 79.042 &  84.109 &  81.521  \\
    LPC & \multirow{3}{*}{{LDC}}  & 87.294 & 86.511 &  88.559 &  86.424 \\ %\cline{2-7}
   
    LDC & & \textbf{94.706} &  \textbf{93.589}  &  \textbf{93.210}  &  \textbf{94.079} \\ %\cline{2-7}
   
    SHC+LPC+LDC & &  87.758 &  86.686  &  87.869  &  91.918\\
\hline 
   
  \end{tabular}
  \caption{Results for the MahaBERT models trained on SHC, LDC, and LPC datasets tested on the test set of other datasets in percentage (\%)}
  \label{tbl:resultCA}
\end{table*}
\begin{figure*}[hbt!]
    \centering
   
    \begin{minipage}{0.48\linewidth}
        \centering
        \includegraphics[scale=0.4]{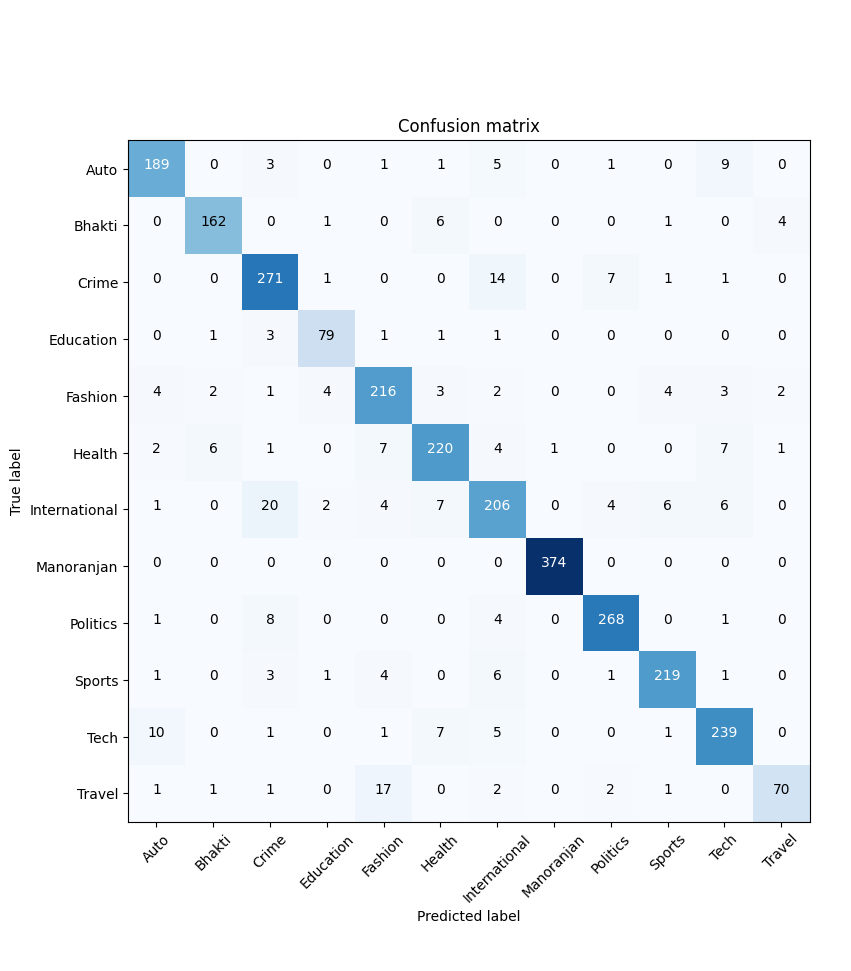}%{c.PNG} % first figure itself
        \caption{Confusion matrix for SHC results}
        \label{fig:cfSHC}
    \end{minipage}\hfill
     \hfill %
    \begin{minipage}{0.48\linewidth}
        \centering
        \includegraphics[scale=0.4]{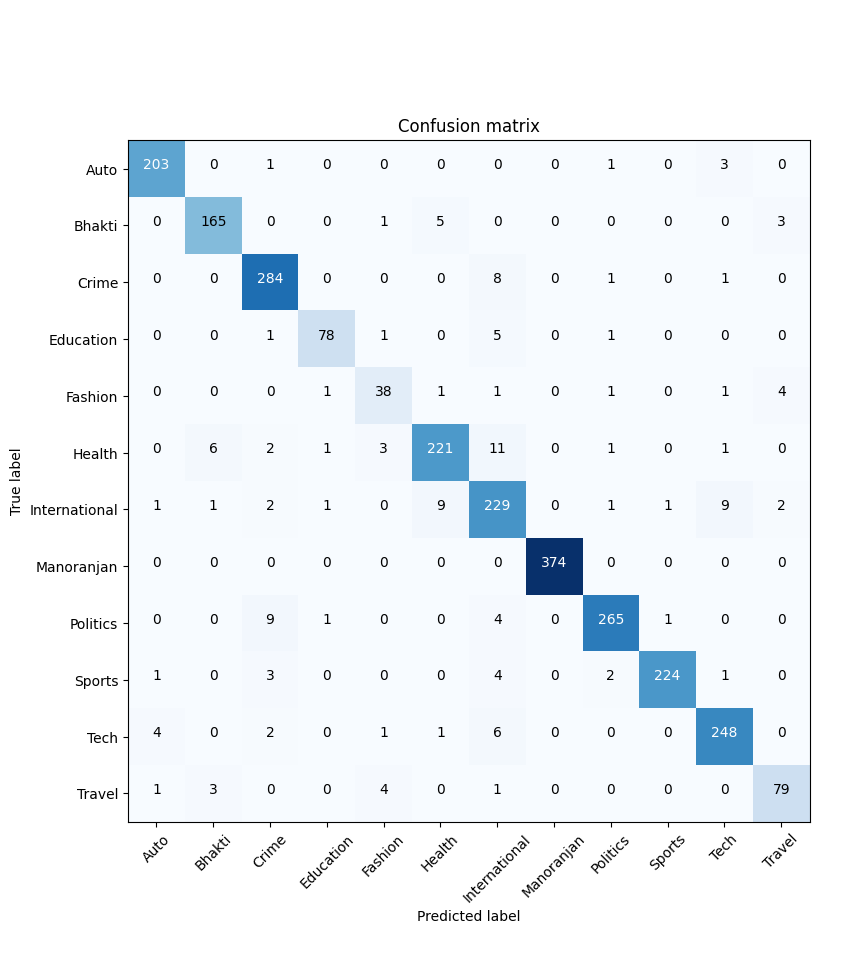}%{h.PNG} % second figure itself
        \caption{Confusion matrix for LDC results}
        \label{fig:cfLDC}
    \end{minipage}
    \includegraphics[scale=0.4]{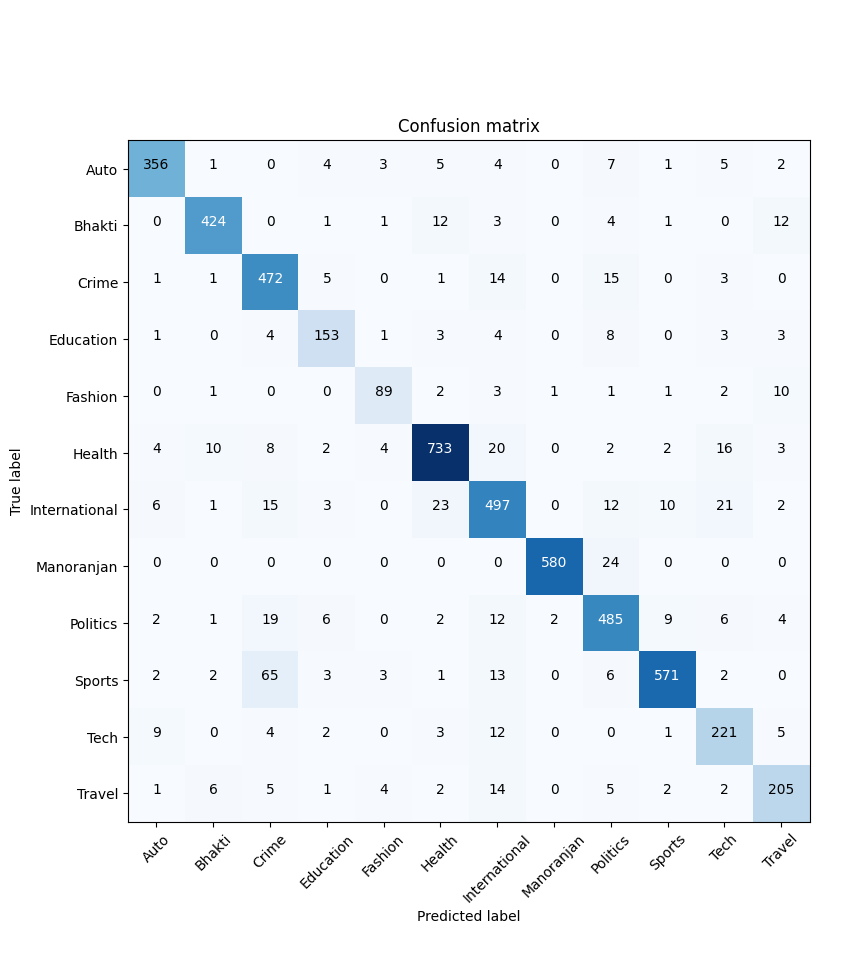}%{c.PNG} % first figure itself
        \caption{Confusion matrix for LPC results}
        \label{fig:cfLPC}
    
\end{figure*}

\section{Evaluation}
We fine-tune the monolingual and multilingual BERT models supporting the Marathi language on the curated L3Cube-MahaNews corpus for the text classification task. A dense layer is added on top of the BERT model which maps the [CLS] token embedding to the 12 target labels.
%The fine-tuned text classifier model when given a text input, classifies it into one of the categorical labels such as Auto, Bhakti, Crime, Politics etc as mentioned above.
\subsection{Experiment setup}
\subsubsection{Data Preparation}
Each of the SHC, LDC, and LPC corpora are split into train, test, and validation datasets in a ratio of 80:10:10. We have ensured that the category-wise distribution ratio of data in SHC, LDC, and LPC remains constant in the split datasets.

% For better understanding, let's assume that the 12 categories are in a ratio of:
% \begin{equation}
% x_1:x_2:x_3:x_4:x_5:x_6:......x_{12}
% \end{equation}

% Then, the training dataset will have the 12 categories in the same ratio:
% \begin{equation}
% 0.8x_1:0.8x_2:0.8x_3:0.8x_4:......0.8x_{12}
% \end{equation}

% Then, the test and validation datasets will have the 12 categories in the same ratio:
% \begin{equation}
% 0.1x_1:0.1x_2:0.1x_3:0.1x_4:......0.1x_{12}
% \end{equation}

The datasets are preprocessed to remove unwanted characters and words from it such as newline characters, hashtags, URLs, and so on. After preprocessing, only Devanagari, English, and numerical digits are retained.

Refer to Table~\ref{tbl:split} for the category-wise distribution of data into train, test, and validation datasets.

% \begin{table*}[htbp]
%   \centering
%   \begin{tabular}{|c|c|c|c|c|c|c|c|c|c|}
%     \hline
%     \multicolumn{1}{|c|}{Category} & \multicolumn{3}{|c|}{SHC} & \multicolumn{3}{|c|}{LDC} & \multicolumn{3}{|c|}{LPC}
% %     \begin{table}[ht]
% % \caption{Multi-column table}

%   %  \hline  
%   \end{tabular}
%   \caption{Results for all the models trained on SHC, LDC and LPC datasets in percentage (\%)}
%   \label{tbl:result}
% \end{table*}

\subsubsection{Models}
The pre-trained BERT models that have been finetuned for text classification are as follows:
\begin{itemize}
    \item \textit{\textbf{MahaBERT}}\footnote{\url{https://huggingface.co/l3cube-pune/marathi-bert-v2}}:
    MahaBERT is a 752
million token  multilingual BERT model fine-tuned on L3Cube-MahaCorpus and other publicly available Marathi monolingual datasets.
    \item \textit{\textbf{indicBERT}}\footnote{\url{https://huggingface.co/ai4bharat/indic-bert}}: IndicBERT is a multi-lingual AlBERT model exclusively pre-trained on 12 Indian languages. It is pre-trained on AI4Bharat IndicNLP Corpora of around 9 billion tokens.
    \item \textit{\textbf{MuRIL}}\footnote{\url{https://huggingface.co/google/muril-base-cased}}:
    MuRIL is a BERT model pre-trained in 17 Indian languages. It has been pre-trained on datasets from Wikipedia, Common Crawl, Dakshina, etc.  
\end{itemize}

It was found that these models gave the best results when they were trained for 3 epochs on the training datasets at the default learning rate (1e-3). The MuRIL model, on the other hand, performed best during 5 training epochs for the SHC dataset.

The fine-tuned MahaBERT models were also tested against test sets of the other datasets like the pre-trained MahaBERT model fine-tuned on SHC dataset was tested against the test sets for LDC and LPC to compute the results of this cross-analysis. 
\subsection{Results}
The results obtained from fine-tuning the models on our datasets are shown in Table~\ref{tbl:result} along with the confusion matrices in Figure~\ref{fig:cfSHC}, \ref{fig:cfLDC} and \ref{fig:cfLPC}.
The results obtained on performing the cross-analysis by testing the MahaBERT model on test sets of different datasets can be referred from Table~\ref{tbl:resultCA}
\\The key observations that were inferred are as follows:
\begin{itemize}
    \item The monolingual MahaBERT model outperforms all other models in terms of the various scores depicted in the table for every corpus.
    \item Among SHC, LDC, and LPC, LDC gave the best results in fine-tuning for the text classification task. This is expected as the long document data contains more information as compared to the other two smaller-length datasets.
    \item LPC reports scores on the lower side for all the 3 models. A paragraph might at times contain more generic information and hence result in confusion for the models.
    \item A cross-dataset testing or zero-shot testing on unseen datasets reveals that models trained on one dataset don't generalize well on other test sets. This affirms the need for different datasets with varying text lengths. 
    \item The model trained on all three datasets (SHC + LPC + LDC) provides the best results for LPC but fares poorly for SHC and LDC. This shows that building a single competitive model needs more attention. More samples in LPC dataset as compared to other datasets could explain the bias towards LPC. An extensive evaluation of this behavior is left to future scope.
\end{itemize}

\section{Conclusion}
In this paper, we present L3Cube-MahaNews - a suite of 3 labeled datasets that consists of 1.08L+ Marathi records for the Marathi Text Classification. The paper describes an extensive set of 12 categorical labels used to create the supervised datasets. We have performed fine-tuning on Marathi-based models to provide a benchmark for future studies and development. The models utilized were MahaBERT, IndicBERT, and MuRIL. We report the best accuracy using MahaBERT for the LDC dataset. We hope that our datasets will play an important role in the betterment of Marathi language support in the field of NLP.
\section*{Limitations}
 During data scrapping and preparation, it was seen that some news articles had scanned images, GIFs, banner ads, etc., as a part of web page content. Thus, additional tools (e.g. OCR-image-to-text converter) might be required to extract text from such web content and retain only the news-related textual data in a proper format. Moreover, since the LPC dataset was created by extracting random paragraphs from the parent articles, these might at times contain generic information not specific to the target label. In future we can manually verify the dataset to filter such problematic entiries.

 \section*{Acknowledgments}
This work was done under the L3Cube Pune mentorship
program. We would like to express our gratitude towards
our mentors at L3Cube for their continuous support and
encouragement. This work is a part of the L3Cube-MahaNLP project \cite{joshi2022l3cube_mahanlp}.
 
% \section*{Acknowledgements}
% This work is a part of the L3Cube Pune mentorship program. We would like to convey our appreciation and gratitude to our L3Cube mentors for providing us with various opportunities and guiding us in the same.
% \section*{Reference}
\interlinepenalty=5000

\bibliographystyle{acl_natbib}
\bibliography{custom}

\end{document}